\documentclass[11pt,a4paper]{article}
\usepackage[hyperref]{acl2021}
\usepackage{times}
\usepackage{latexsym}

\usepackage{microtype}

\aclfinalcopy 

\usepackage{booktabs, tabularx}
\usepackage{graphics}
\usepackage{multirow}
\usepackage{amsmath}
\usepackage{array,multirow,graphicx}

\usepackage{CJKutf8}

\newcommand{\cmt}[1]{\ignorespaces}

\usepackage{amsmath}
\usepackage{multirow}
\usepackage{booktabs}
\usepackage{tikz-dependency}
\usepackage{pgfplots}
\usepackage{dashrule}
\usepackage{booktabs,arydshln}
\usepackage{scalerel}
\definecolor{skyblue}{RGB}{70, 130, 180}
\definecolor{fontgray}{RGB}{44, 62, 80}
\definecolor{myred}{RGB}{235, 47, 6} 
\definecolor{myblue}{RGB}{0, 168, 255}
\definecolor{lightorange}{RGB}{244, 122, 96}
\definecolor{dsblue}{RGB}{65,105,225}
\definecolor{spgreen}{RGB}{46,139,87}
\definecolor{tored}{RGB}{220,20,60}
\definecolor{glyellow}{RGB}{218,165,32}
\definecolor{botcolor}{RGB}{119,136,153}
\definecolor{linecolor}{RGB}{128,128,128}
\definecolor{topcolor}{RGB}{0,0,139}
\definecolor{nblue}{RGB}{0, 114, 189}
\definecolor{norange}{RGB}{217, 83, 25}
\definecolor{nyellow}{RGB}{237, 177, 32}
\definecolor{npurple}{RGB}{126, 47, 142}
\definecolor{ngreen}{RGB}{119, 172, 48}
\definecolor{nlightblue}{RGB}{77, 190, 238}
\definecolor{nred}{RGB}{162, 20, 47}
\definecolor{lowgreen}{RGB}{186, 220, 88} 
\definecolor{littledarkblue}{RGB}{48, 51, 107}
\definecolor{mygray}{RGB}{158,158,158}
\definecolor{fruitpurple}{RGB}{108, 92, 231}
\definecolor{river}{RGB}{52,152,219}
\definecolor{purple}{RGB}{142, 68, 173} 

\newcommand*{\affaddr}[1]{#1} 
\newcommand*{\affmark}[1][*]{\textsuperscript{#1}}

\title{Learning Span-Level Interactions for Aspect Sentiment Triplet Extraction}

\author{%
Lu Xu\affmark[* 1, 2]\thanks{$*$ Equal contribution.  Lu Xu and Yew Ken Chia are under the Joint PhD Program between Alibaba and Singapore University of Technology and Design. },
Yew Ken Chia\affmark[* 1, 2], Lidong Bing\affmark[2]\\
\affaddr{\affmark[1] Singapore University of Technology and Design}\\
\affaddr{\affmark[2] DAMO Academy, Alibaba Group}\\
\tt{ \{xu\_lu,yewken\_chia\}@mymail.sutd.edu.sg}\\
\tt{l.bing@alibaba-inc.com}\\
}

\renewcommand\footnotemark{}

\date{}

\begin{document}
\maketitle
\begin{abstract} 
Aspect Sentiment Triplet Extraction (ASTE) is the most recent subtask of ABSA which outputs triplets of an aspect target, its associated sentiment, and the corresponding opinion term.
Recent models perform the triplet extraction in an end-to-end manner but heavily rely on the interactions 
between each target word and opinion word.
Thereby, they cannot perform well on targets and opinions which contain multiple words.
Our proposed span-level approach explicitly considers the interaction between the whole spans of targets and opinions when predicting their sentiment relation.
Thus, it can make predictions with the semantics of whole spans, ensuring better sentiment consistency.
To ease the high computational cost caused by span enumeration, we propose a dual-channel span pruning strategy by incorporating supervision from the Aspect Term Extraction (ATE) and Opinion Term Extraction (OTE) tasks.
This strategy not only improves computational efficiency but also distinguishes the opinion and target spans more properly.
Our framework simultaneously achieves strong performance for the ASTE as well as ATE and OTE tasks.
In particular, our analysis shows that our span-level approach achieves more significant improvements over the baselines on triplets with multi-word targets or opinions.
\footnote{We make our code publicly available at \url{https://github.com/chiayewken/Span-ASTE}.}

\end{abstract}

\section{Introduction}
\label{sec:intro}
\begin{figure}[!t]
\centering
\includegraphics[width=1.0\columnwidth]{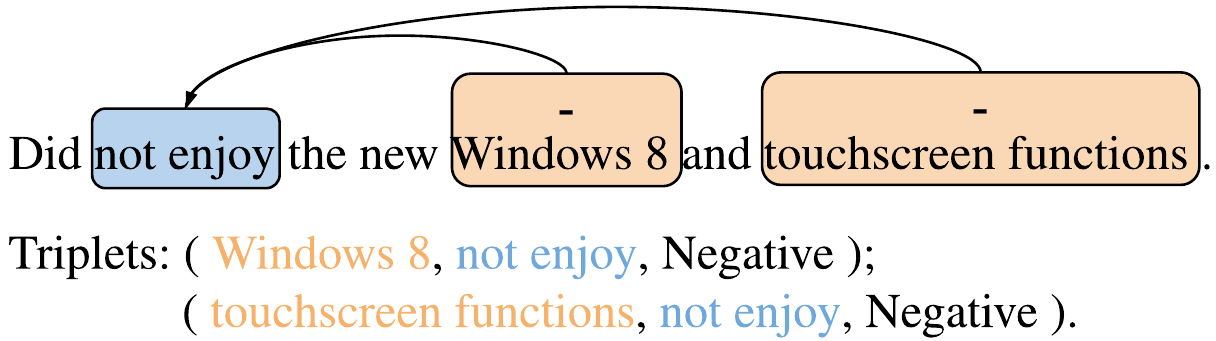}
\caption{An example of ASTE. The spans highlighted in orange are target terms, and the span in blue is opinion term. The ``-'' on top of target terms indicates negative sentiment.}
\label{fig:example}
\end{figure}

Aspect-Based Sentiment Analysis (ABSA) \cite{liu2012sentiment, pontiki-EtAl:2014:SemEval} is an aggregation of several fine-grained sentiment analysis tasks, 
and its various subtasks 
are designed with the aspect target as the fundamental item.
For the example in Figure \ref{fig:example}, the aspect targets are \textit{``Windows 8''} and \textit{``touchscreen functions''}.
Aspect Sentiment Classification (ASC) \cite{dong2014adaptive, zhang2016gated, Yang2017AttentionBL, tnet2018, Tang:ACL2019} is one of the most well-explored subtasks of ABSA and aims to predict the sentiment polarity of a given aspect target.
However, it is not always practical to assume that the aspect target is provided.
Aspect Term Extraction (ATE) \cite{yin2016unsupervised, li2018aspect, ma-etal-2019-exploring} focuses on extracting aspect targets, while Opinion Term Extraction (OTE) \cite{yang-cardie-2012-extracting, kinger2013, yang-cardie-2013-joint}
aims to extract the opinion terms which largely determine the sentiment polarity of the sentence or the corresponding target term.
Aspect Sentiment Triplet Extraction (ASTE) \cite{peng2019knowing} is the most recently proposed subtask of ABSA, which forms 
a more complete picture of the sentiment information through the triplet of an aspect target term, the corresponding opinion term, and the expressed sentiment.
For the example in Figure \ref{fig:example}, there are two triplets: (\textit{``Windows 8''},  \textit{``not enjoy''}, Negative) and (\textit{``touchscreen functions''}, \textit{``not enjoy''}, Negative).

The initial approach to ASTE \cite{peng2019knowing} was a two-stage pipeline.
The first stage extracts target terms and their sentiments via a joint labeling scheme 
\footnote{For example, the joint tag ``B-POS'' denotes the beginning of a target span with positive sentiment polarity.},
as well as the opinion terms with standard BIOES \footnote{A common tagging scheme for sequence labeling, denoting “begin, inside, outside, end and single” respectively.}
tags.
The second stage then couples the extracted target and opinion terms to determine their paired sentiment relation.
We know that in ABSA, the aspect sentiment is mostly determined by the opinion terms expressed on the aspect target \cite{qiu-etal-2011-opinion, yang-cardie-2012-extracting}. 
However, this pipeline approach breaks the interaction within the triplet structure. 
Moreover, pipeline approaches usually suffer from the error propagation problem.

Recent end-to-end approaches \cite{wu-etal-2020-grid,Xu2020PositionAwareTF,zhang-etal-2020-multi-task}
can jointly extract the target and opinion terms and classify their sentiment relation. One drawback is that they heavily rely on word-to-word interactions to predict the sentiment relation for the target-opinion pair.
Note that it is common for the aspect targets and opinions to contain multiple words, which accounts for roughly one-third of triplets in the benchmark datasets. 
However, the previous methods \cite{wu-etal-2020-grid,zhang-etal-2020-multi-task}
predict the sentiment polarity for each word-word pair independently, which cannot guarantee their sentiment consistency when forming a triplet.
As a result, this prediction limitation on triplets that contain multi-word targets or opinions
inevitably hurts the overall ASTE performance.
For the example in Figure~\ref{fig:example}, by only considering the word-to-word interactions, it is easy to wrongly predict that 
\textit{``enjoy''} expresses a positive sentiment on \textit{``Windows''}.
\citet{Xu2020PositionAwareTF} proposed a position-aware tagging scheme to allow the model to couple each word in a target span with all possible opinion spans,
i.e., aspect word to opinion span interactions (or vice versa, aspect span to opinion word interactions).
However, it still cannot directly model the span-to-span interactions between the whole target spans and opinion spans. 


In this paper, we propose a span-based model for ASTE (Span-ASTE), which for the first time directly captures the span-to-span interactions when predicting the sentiment relation of an aspect target and opinion pair.
Of course, it can also consider the single-word aspects or opinions properly. 
Our model explicitly generates span representations for all possible target and opinion spans,
and their paired sentiment relation is independently predicted for all possible target and opinion pairs.
Span-based methods have shown encouraging performance on other tasks, such as coreference resolution \cite{lee-etal-2017-end}, semantic role labeling \cite{he-etal-2018-jointly}, and relation extraction \cite{luan-etal-2019-general, Wadden2019EntityRA}.
However, they cannot be directly applied to the ASTE task due to different task-specific characteristics.


Our contribution can be summarized as follows: 
\begin{itemize}
\item We tailor a span-level approach to explicitly consider the span-to-span interactions for the ASTE task and conduct extensive analysis to demonstrate its effectiveness. Our approach significantly improves performance, especially on triplets which contain multi-word targets or opinions.
\item We propose a dual-channel span pruning strategy by incorporating explicit supervision from the ATE and OTE tasks to ease the high computational cost caused by span enumeration and maximize the chances of pairing valid target and opinion candidates together.
\item Our proposed Span-ASTE model outperforms the previous methods significantly not only for the ASTE task, but also for the ATE and OTE tasks on four benchmark datasets with both BiLSTM and BERT encoders.
    
\end{itemize}

\section{Span-based ASTE}

\begin{figure*}[!t]
\centering
\includegraphics[width=1.0\textwidth]{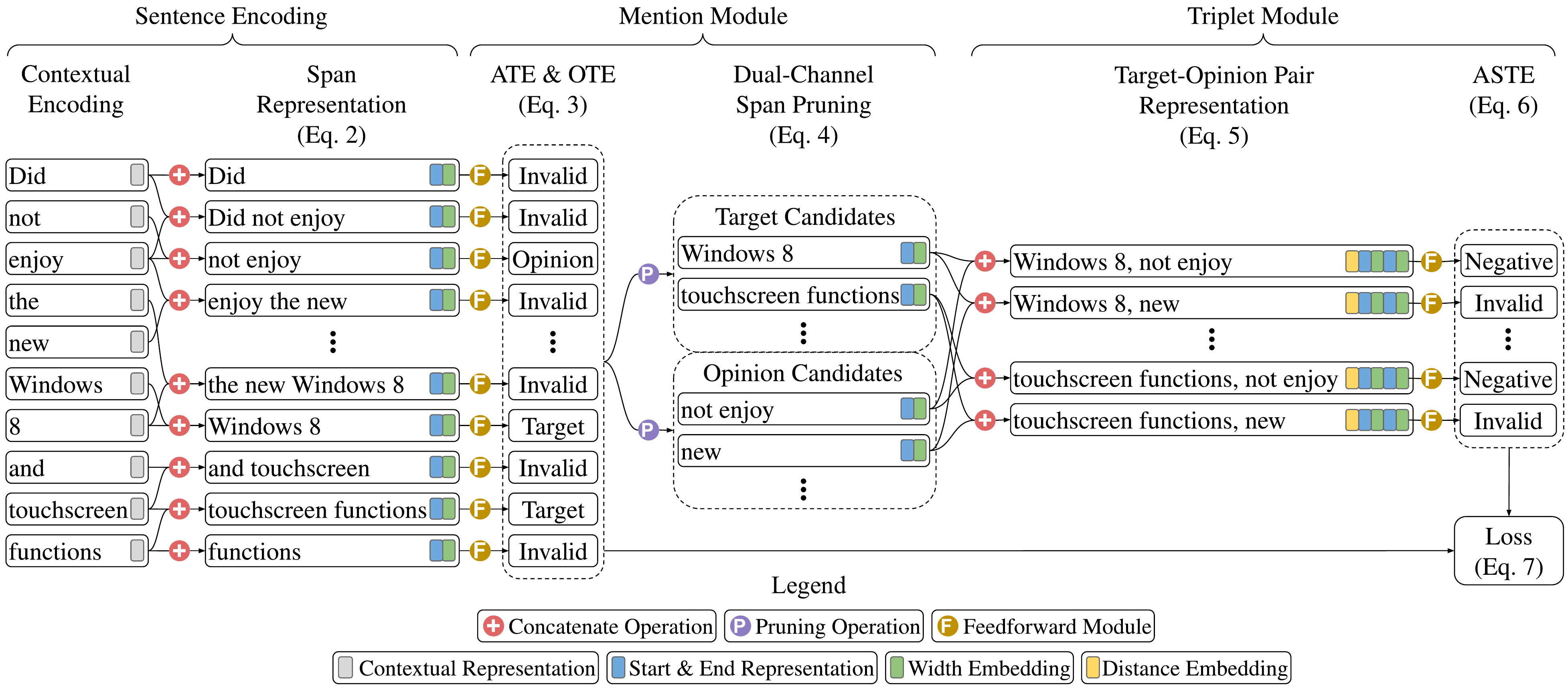}
\caption{Span-ASTE model structure.}
\label{fig:model}
\end{figure*}

\subsection{{Task Formulation}}
Let $X = \{x_1, \; x_2, \; ..., \; x_n\}$ denote a sentence of $n$ tokens, let $S = \{s_{1,1}, \; s_{1,2}, \; ..., \; s_{i,j}, \; ..., \; s_{n,n}\}$ be the set of all possible \textcolor{black}{enumerated} spans in $X$, with $i$ and $j$ indicating the start and end positions of a span in the sentence.
\textcolor{black}{We limit the span length as $0 \leq j-i \leq L$.}
The objective of the ASTE task is to extract all possible triplets in $X$.
\textcolor{black}{Each sentiment triplet is defined as $(target, opinion, sentiment)$ where $sentiment \in \{Positive, Negative, Neutral\}$.}

\subsection{Model Architecture}
As shown in Figure \ref{fig:model}, 
Span-ASTE consists of three modules: sentence encoding, mention module, and triplet module. 
For the given example, the sentence is first input to the sentence encoding module to obtain the token-level representation, from which we derive the span-level representation for each enumerated span, such as \textit{``did not enjoy''}, \textit{``Windows 8''}. 
We then adopt the ATE and OTE tasks to supervise our proposed dual-channel span pruning strategy which obtains the pruned target and opinion candidates, such as \textit{``Windows 8''} and \textit{``not enjoy''} respectively.
Finally, each target candidate and opinion candidate are coupled to determine the sentiment relation between them.


\subsubsection{Sentence Encoding}
We explore two encoding methods to obtain the contextualized representation for each word in a sentence: BiLSTM and BERT.
\paragraph{BiLSTM}
We 
first obtain the word representations $\{\mathbf{e_1}, \mathbf{e_2}, ...,\mathbf{e_i}, ..., \mathbf{e_n}\}$ from the 
\textcolor{black}{300-dimension pre-trained GloVe \cite{pennington2014glove} \cmt{word} embeddings}
which are then contextualized by a bidirectional LSTM \cite{lstm97} layer.
The $i^{th}$ token is represented as:
\begin{equation}
    \mathbf{h}_i=[\overrightarrow{\mathbf{h}_i}; \; \overleftarrow{\mathbf{h}_i}]
\end{equation}
where $\overrightarrow{\mathbf{h}_i}$ and $\overleftarrow{\mathbf{h}_i}$ are the hidden states of the forward and backward LSTMs respectively. 

\paragraph{BERT} 
An alternative encoding method 
is to use a pre-trained language model such as BERT \cite{devlin2019bert} to obtain the contextualized word representations
$\mathbf{x}= [ \mathbf{x}_1, \mathbf{x}_2, ..., \mathbf{x}_n]$.
For words that are tokenized as multiple word pieces, we use mean pooling to aggregate their representations \cmt{across word pieces}.

\paragraph{Span Representation}
\label{representation}
We define each span representation $s_{i,j} \in S$ as:
\begin{equation}
\label{spanrep}
    \mathbf{s}_{i,j} =
    \begin{cases} 
        [\mathbf{h}_i ; \; \mathbf{h}_j ;\;
        f_{width}(i,j)]  & \text{if BiLSTM}
        \\
        [\mathbf{x}_i ; \; \mathbf{x}_j ;\;
        f_{width}(i,j)]  & \text{if BERT}
    \end{cases}
\end{equation}
where $f_{width}(i,j)$ produces a trainable feature embedding representing the span width (i.e., $j - i + 1$).
Besides the concatenation of the start token, end token, and width representations,
the span representation $\mathbf{s}_{i,j}$ can also be formed by max-pooling or mean-pooling across all token representations of the span from position $i$ to $j$.  The experimental results can be found in the ablation study.


\subsubsection{Mention Module}
\label{mention_module}
\paragraph{ATE \& OTE Tasks}
We employ the ABSA subtasks of ATE and OTE to 
guide our dual-channel span pruning strategy through the scores of the predicted opinion and target span.
Note that the target terms and opinion terms are not yet paired together at this stage.  
The mention module takes the representation of each enumerated span  $\mathbf{s}_{i, j} $ as input and predicts the mention types $ m \in \{Target, Opinion, Invalid\} $.
\begin{equation}
\label{aux_task}
P(m|\mathbf{s}_{i, j}) = \mathrm{softmax}(\mathrm{FFNN}_m (\mathbf{s}_{i, j}))
\end{equation}
where $\mathrm{FFNN}$ denotes a feed-forward neural network with non-linear activation.

\begin{table*}[t]
\centering
\resizebox{1\linewidth}{!}{
\begin{tabular}{@{~}l@{~}r@{~}@{~}r@{~}@{~}r@{~}@{~}r@{~}@{~}r@{~}@{~}r@{~}@{~}r@{~}@{~}r@{~}@{~}r@{~}@{~}r@{~}@{~}r@{~}@{~}r@{~}@{~}r@{~}@{~}r@{~}@{~}r@{~}@{~}r@{~}@{~}r@{~}@{~}r@{~}@{~}r@{~}@{~}r@{~}@{~}r@{~}@{~}r@{~}@{~}r@{~}@{~}r@{~}}
\toprule 
\multirow{2}{*}{\textbf{Dataset}} & \multicolumn{6}{c}{\textbf{Rest 14}} & \multicolumn{6}{c}{\textbf{Lap 14}} & \multicolumn{6}{c}{\textbf{Rest 15}} & \multicolumn{6}{c}{\textbf{Rest 16}} \\  \cmidrule(lr){2-7}\cmidrule(lr){8-13}\cmidrule(lr){14-19}\cmidrule(lr){20-25}
& \#S{\color{white},} & \# +{\color{white},} & \# 0{\color{white},} & \# -{\color{white},} & \#SW & \#MW&  \#S{\color{white},} & \# +{\color{white},} & \# 0{\color{white},} & \# -{\color{white},} & \#SW & \#MW&\#S{\color{white},} & \# +{\color{white},} & \# 0{\color{white},} & \# -{\color{white},} & \#SW & \#MW&\#S{\color{white},} & \# +{\color{white},} & \# 0{\color{white},} & \# -{\color{white},} & \#SW & \#MW\\\midrule
\textbf{Train} & 1266 & 1692 & 166 & 480 & 1586 & 752 &
906 & 817 & 126 & 517 & 824 & 636 &
605 & 783 & 25 & 205 & 678 & 335 &
857 &  1015 & 50 & 329 & 918 & 476 \\ 
\textbf{Dev} & 310 & 404 & 54 & 119  & 388 & 189 &
219 &169 & 36 & 141 & 190 & 156 &
148 &185 & 11 & 53 & 165 & 84 &
210 &252 & 11 & 76 & 216 & 123 \\
\textbf{Test} & 492 & 773 & 66 & 155 & 657 & 337 &
328 & 364 & 63 & 116 & 291 & 252 &
322 & 317 & 25 & 143 & 297 & 188 &
326 & 407 & 29 & 78 & 344 & 170  \\
\bottomrule
\end{tabular}
}
\caption{Statistics of datasets. \#S denotes the number of sentences.
\# +, \# 0, and \# - denote the numbers of positive, neutral, and negative sentiment triplets respectively.
\#SW denotes the number of triplets where both target and opinion terms are single-word spans. 
\#MW denotes the number of triplets where at least one of the target or opinion terms are multi-word spans. 
}
\label{tab:dataset}
\end{table*}

\paragraph{Pruned Target and Opinion} 

For a sentence X of length $n$, the number of enumerated spans is $O(n^2)$,
while the number of possible pairs between all opinion and target candidate spans is $O(n^4)$ at the later stage (i.e., the triplet module).
As such, it is not computationally practical to consider all possible pairwise interactions when using a span-based approach.
Previous works \cite{luan-etal-2019-general, Wadden2019EntityRA} employ a pruning strategy to reduce the number of spans, but they only prune the spans to a single pool which is a mix of different mention types. 
This strategy does not fully consider the structure of an aspect sentiment triplet as it does not recognize the fundamental difference between a target and an opinion term.
Hence, we propose to use a dual-channel pruning strategy which results in two separate pruned pools of aspects and opinions. This minimizes computational costs while maximizing the chance of pairing valid opinion and target spans together.
 The opinion and target candidates are selected based on the scores of the mention types for each span based on Equation \ref{aux_task}: 
\begin{equation}
\begin{split}
&\Phi_{target}(\mathbf{s}_{i,j}) = P(m=target|\mathbf{s}_{i, j}) \\
&\Phi_{opinion}(\mathbf{s}_{i,j}) = P(m=opinion|\mathbf{s}_{i, j})
\end{split}
\end{equation} 

We use the mention scores $\Phi_{target}$ and $\Phi_{opinion}$ 
to select the top candidates from the enumerated spans 
and obtain the target candidate pool $S^t=\{ ...,  \mathbf{s}^t_{a,b}, ... \}$
and the opinion candidate pool  $S^o=\{ ...,  \mathbf{s}^o_{c,d}, ... \}$ respectively.
To consider a proportionate number of candidates for each sentence,
the number of selected spans for both pruned target and opinion candidates is $nz$, where $n$ is the sentence length and $z$ is a threshold hyper-parameter.
Note that although the pruning operation 
prevents the gradient flow back to the FFNN in the mention module,
it is already receiving supervision from the ATE and OTE tasks.
Hence, our model can be trained end-to-end without any issue or instability.


\subsubsection{Triplet Module}
\label{triplet_module}
\paragraph{Target Opinion Pair Representation}

We obtain the target-opinion pair representation by coupling each target candidate representation $\mathbf{s}_{a,b}^t \in S^t$ with each opinion candidate representation $\mathbf{s}_{c,d}^o \in S^o$:
\begin{equation}
\label{pair_concat}
\mathbf{g}_{\mathbf{s}_{a,b}^t, \mathbf{s}^o_{c,d}} = [\mathbf{s}_{a,b}^t; \;\mathbf{s}^o_{c,d};
\;f_{distance}(a, b, c, d)
]
\end{equation} 
where $f_{distance}(a, b, c, d)$ produces a trainable feature embedding based on the distance (i.e., $min(|b-c|, |a-d|)$) between the target and opinion spans, following \cite{lee-etal-2017-end, he-etal-2018-jointly,Xu2020PositionAwareTF}.


\paragraph{Sentiment Relation Classifier}
Then, we input the span pair representation $\mathbf{g}_{\mathbf{s}_{a,b}^t, \mathbf{s}^o_{c,d}}$ to a feed-forward neural network to determine the probability of sentiment relation $r \in R=\{Positive, Negative, Neutral, Invalid\}$ between the target $\mathbf{s}_{a,b}^t$ and the opinion $\mathbf{s}_{c,d}^o$:
\begin{equation}
\label{pair_classifier}
    P(r|\mathbf{s}_{a,b}^t, \mathbf{s}^o_{c,d}) = 
\mathrm{softmax}(\mathrm{FFNN}_{r} (\mathbf{g}_{\mathbf{s}_{a,b}^t, \mathbf{s}^o_{c,d}}))
\end{equation}
$Invalid$ here indicates that the target and opinion pair has no valid sentiment relationship.

\subsection{Training}
\label{loss}
The training objective is defined as the sum of the negative log-likelihood from both the mention module and triplet module.
\begin{equation}
\begin{split}
    \mathcal{L} =
    &- \sum_{\mathbf{s}_{i,j} \in S}\log P(m_{i,j}^* | \mathbf{s}_{i,j}) \\
    &-\sum_{\mathbf{s}_{a,b}^t \in S^t, \mathbf{s}^o_{c,d} \in S^o}\log P(r^*|\mathbf{s}_{a,b}^t, \mathbf{s}^o_{c,d})
\end{split}
\end{equation}
where $m_{i,j}^*$ is the gold mention type of the span $\mathbf{s}_{i,j}$, and $r^*$ is the gold sentiment relation of the target and opinion span pair ($\mathbf{s}_{a,b}^t, \mathbf{s}^o_{c,d}$).  $S$ indicates the enumerated span pool; $S^t$ and $S^o$ are the pruned target and opinion span candidates. 

\begin{table*}[!t]
    \centering
    \resizebox{1\textwidth}{!}{%
    \begin{tabular}{llcccccccccccc}
    \toprule
    & \multirow{2}{*}{\textbf{Model}}  & \multicolumn{3}{c}{ \textbf{Rest 14}} & \multicolumn{3}{c}{ \textbf{ Lap 14}} & \multicolumn{3}{c}{ \textbf{Rest 15}}  & \multicolumn{3}{c}{  \textbf{Rest 16}}    \\ \cmidrule(lr){3-5} \cmidrule(lr){6-8} \cmidrule(lr){9-11} \cmidrule(lr){12-14}
    && $P.$ & $R.$ & $F_1$& $P.$ & $R.$ & $F_1$& $P.$ & $R.$ & $F_1$ & $P.$ & $R.$ & $F_1$ \\
    \midrule
    \multirow{7}{*}{\rotatebox[origin=c]{90}{\textbf{BiLSTM}} } &{CMLA+} \cite{wang2017coupled}$^{\dag}$&39.18 & 47.13 & 42.79 &  30.09 & 36.92 & 33.16 & 34.56 & 39.84 & 37.01 & 41.34 & 42.10 & 41.72          \\
     &{RINANTE+} \cite{dai2019neural}$^{\dag}$&  31.42 & 39.38 & 34.95 &  21.71 & 18.66 & 20.07  & 29.88 & 30.06 & 29.97 & 25.68 & 22.30 & 23.87  \\
     &{Li-unified-R} \cite{li2019unified}$^{\dag}$& 41.04 & 67.35 & 51.00 &  40.56 & 44.28 & 42.34 & 44.72 & 51.39 & 47.82 &  37.33 & 54.51 & 44.31 \\
     &\citet{peng2019knowing}$^{\dag}$ & 43.24 & 63.66 & 51.46 & 37.38 & 50.38 & 42.87 & 48.07 & 57.51 & 52.32 & 46.96 & 64.24 & 54.21 \\ 
     \cmidrule(lr){2-14}
     &\citet{zhang-etal-2020-multi-task} $^*$ &62.70 & 57.10 & 59.71 & 49.62 & 41.07 & 44.78 & 55.63 & 42.51 & 47.94 & 60.95 & 53.35 & 56.82 \\
    &{GTS} \cite{wu-etal-2020-grid}$^*$ & 66.13 & 57.91 & 61.73 & 53.35 & 40.99 & 46.31 & 60.10 & 46.89 & 52.66 & 63.28 & 58.56 & 60.79 \\

     &{JET}$^o_{M=6}$ \cite{Xu2020PositionAwareTF}$^{\dag}$ & 61.50 &55.13& 58.14& 53.03 &33.89&41.35 &64.37&44.33&	52.50 & 70.94 & 57.00 & 63.21\\
     \cmidrule(lr){2-14}
     &\textbf{Span-ASTE} (Ours)  & 72.52 & 62.43 & \textbf{67.08} & 59.85 & 45.67 & \textbf{51.80} & 64.29 & 52.12 & \textbf{57.56} & 67.25 & 61.75 & \textbf{64.37} \\

    \midrule
    \multirow{3}{*}{\rotatebox[origin=c]{90}{\textbf{BERT}} }
    &{GTS} \cite{wu-etal-2020-grid}$^*$ & 67.76 & 67.29 & 67.50 &57.82 & 51.32 & 54.36 & 62.59 & 57.94 & 60.15 & 66.08	& 69.91	& 67.93 \\

     &{JET}$^o_{M=6}$ \cite{Xu2020PositionAwareTF} $^{\dag}$& 70.56 & 55.94 & 62.40 & 55.39 & 47.33 & 51.04 & 64.45 & 51.96 & 57.53 & 70.42 & 58.37 & 63.83 \\
    \cmidrule(lr){2-14}
     &\textbf{Span-ASTE} (Ours) &  72.89 & 70.89 & \textbf{71.85} & 63.44 & 55.84 & \textbf{59.38} & 62.18 & 64.45 & \textbf{63.27} & 69.45 & 71.17 & \textbf{70.26} \\

  \bottomrule
    \end{tabular}
    }
    \caption{
    Results on the test set of the ASTE task. 
    $^{\dag}$: The results are retrieved from \citet{Xu2020PositionAwareTF}.
    $^*$: For a fair comparison, we reproduce the results using their released implementation code and configuration on the same ASTE datasets released by \citet{Xu2020PositionAwareTF}.
    } 
    \label{tab:main_results}
\end{table*}

\section{Experiment}
\subsection{Datasets}

Our proposed Span-ASTE model is evaluated on four ASTE datasets released by \citet{Xu2020PositionAwareTF}, which include three datasets in the restaurant domain and one dataset in the laptop domain. The first version of the ASTE datasets are released by \citet{peng2019knowing}. However, it is found that not all triplets are explicitly annotated \cite{Xu2020PositionAwareTF, wu-etal-2020-grid}.  \citet{Xu2020PositionAwareTF} refined the datasets with the missing triplets and removed triplets with conflicting sentiments.  Note that these four benchmark datasets are derived from the SemEval Challenge \cite{pontiki-EtAl:2014:SemEval, pontiki-etal-2015-semeval, pontiki2016semeval}, and the opinion terms are retrieved from \cite{fan2019target}. Table \ref{tab:dataset} shows the detailed statistics.

\subsection{Experiment Settings}
When using the BiLSTM encoder, the pre-trained GloVe word embeddings are trainable.
The hidden size of the BiLSTM encoder is 300 and the dropout rate is 0.5.
In the second setting, we fine-tune the pre-trained BERT \cite{devlin2019bert} to encode each sentence.
Specifically, we use the 
{uncased version of BERT$_{\text{base}}$}.
The 
model is trained for 10 epochs with a linear warmup for 10\% of the training steps followed by a linear decay of the learning rate to 0.
We employ AdamW as the optimizer with the maximum learning rate of 5e-5 for transformer weights and weight decay of 1e-2. For other parameter groups, we use a learning rate of 1e-3 with no weight decay. 
The maximum span length $L$ is set as 8.
The span pruning threshold $z$ is set as 0.5.
We select the best model 
weights based on the $F_1$ scores on the development set and the reported results are the average of 5 runs with different random seeds.
\footnote{See Appendix for more experimental settings, and also the dev results on the four datasets.}

\subsection{Baselines}

The baselines can be summarized as two groups: pipeline methods and end-to-end methods.

\paragraph{Pipeline} 

For the pipeline approaches listed below, they are modified by \citet{peng2019knowing} to extract the aspect terms together with their associated sentiments via a joint labeling scheme, and opinion terms with BIOES tags at the first stage. At the second stage, the extracted targets and opinions are then paired to determine if they can form a valid triplet. Note that these approaches employ different methods to obtain the features for the first stage.
{CMLA+} \cite{wang2017coupled} employs an attention mechanism to consider the interaction between aspect terms and opinion terms.
{RINANTE+} \cite{dai2019neural} adopts a BiLSTM-CRF model with mined rules to capture the dependency relations.
{Li-unified-R} \cite{li2019unified} uses a unified tagging scheme to 
jointly extract the aspect term and associated sentiment.
\citet{peng2019knowing} includes dependency relation information when considering the interaction between the aspect and opinion terms. 

\paragraph{End-to-end} 
The end-to-end methods aim to jointly extract full triplets in a single stage.  
Previous work by \citet{zhang-etal-2020-multi-task} and \citet{wu-etal-2020-grid} independently predict the sentiment relation for all possible word-word pairs, hence they require decoding heuristics to determine the overall sentiment polarity of a triplet. 
JET \cite{Xu2020PositionAwareTF} models the ASTE task as a structured prediction problem with a position-aware tagging scheme to capture the interaction of the three elements in a triplet.

\subsection{Experiment Results}
Table \ref{tab:main_results} 
compares Span-ASTE with previous models in terms of Precision ($P.$), Recall ($R.$), and $F_1$ scores on four datasets. 
Under the $F_1$ metric, our model consistently outperforms the previous works for 
both BiLSTM and BERT sentence encoders.
In most cases, our model significantly outperforms other end-to-end methods in both precision and recall.
We also observe that
the two strong pipeline methods \cite{li2019unified,peng2019knowing} achieved competitive recall results, but
their overall performance is much worse due to the low precision.
Specifically, using the BiLSTM encoder with GloVe embedding, our model outperforms the best pipeline model \cite{peng2019knowing} by 15.62, 8.93, 5.24, and 10.16 $F_1$ points on the four datasets. 
This result indicates that our end-to-end approach can effectively encode the interaction between target and opinion spans, and also alleviates the error propagation. 
In general, the other end-to-end methods are also more competitive than the pipeline methods.
However, due to the limitations of relying on word-level interactions, their performances are less encouraging in a few cases, such as the results on Lap 14 and Rest 15.
With the BERT encoder, all three end-to-end models
achieve much stronger performance than their LSTM-based versions, which 
is consistent with previous findings 
\cite{devlin2019bert}.
Our approach outperforms the previous best results GTS \cite{wu-etal-2020-grid} by 4.35, 5.02,	3.12, and	2.33 $F_1$ points on the four datasets.  


\subsection{Additional Experiments}
As mentioned in Section \ref{mention_module}, we employ the ABSA subtasks of ATE and OTE to guide our span pruning strategy. 
To examine if 
Span-ASTE can effectively extract target spans and opinion spans, we also evaluate our model on the ATE and OTE tasks on the four datasets.
Table \ref{tab:aote} shows the comparisons of our approach and the previous method GTS \cite{wu-etal-2020-grid}. \footnote{See Appendix for the target and opinion data statistics.
Note that the JET model \cite{Xu2020PositionAwareTF} is not able to directly solve the ATE and OTE tasks unless the evaluation is conducted based on the triplet predictions.
We include such comparisons in the Appendix.}
Without additional retraining or tuning,
our model can directly address the ATE and OTE tasks, with significant performance improvement than GTS in terms of $F_1$ scores on both tasks.
Even though GTS shows a better recall score on the Rest 16 dataset, the low precision score results in 
worse $F_1$ performance.
The better overall performance indicates that our span-level method not only benefits the sentiment triplet extraction,
but also improves the extraction of target and opinion terms by considering the semantics of each whole span rather than relying on decoding heuristics of tagging-based methods.

\begin{table}[!t]
    \centering
    \resizebox{0.999\columnwidth}{!}{%
    \begin{tabular}{llcccccc}
    \toprule
    \multirow{2}{*}{\textbf{Dataset}}  & \multirow{2}{*}{\textbf{Model}}  & \multicolumn{3}{c}{\textbf{ATE}} & \multicolumn{3}{c}{\textbf{OTE}}   \\ \cmidrule(lr){3-5} \cmidrule(lr){6-8} 
    && $P.$ & $R.$ & $F_1$&  $P.$ & $R.$ & $F_1$\\
    \midrule
    \multirow{2}{*}{Rest 14}
    &{GTS}  & 78.12 & 85.64 & 81.69 & 81.12 & 88.24 & 84.53 \\
    &{Ours} & 83.56 & 87.59 & \textbf{85.50} & 82.93 & 89.67 & \textbf{86.16} \\
    
    \midrule
    \multirow{2}{*}{Lap 14}
    &{GTS}  & 76.63 & 82.68 & 79.53 & 76.11 & 78.44 & 77.25 \\
    &{Ours} &  81.48 & 86.39 & \textbf{83.86} & 83.00 & 82.28 & \textbf{82.63 }\\
    
    \midrule
    \multirow{2}{*}{Rest 15}
    &{GTS}  & 75.13 & 81.57 & 78.21 & 74.96 & 82.52 & 78.49 \\
    &{Ours} &  78.97 & 84.68 & \textbf{81.72} & 77.36 & 84.86 & \textbf{80.93}\\
    
    \midrule
    \multirow{2}{*}{Rest 16}
    &{GTS}  & 75.06 & 89.42 & 81.61 & 78.99 & 88.71 & 83.57 \\
    &{Ours} &  79.78 & 88.50 & \textbf{83.89} & 82.59 & 90.91 & \textbf{86.54} \\
 
  \bottomrule
    \end{tabular}
    }
    \caption{Test results on the ATE and OTE tasks with BERT encoder.  
    For reference, we include the results  of the RACL framework \cite{chen-qian-2020-relation} in the Appendix.
    RACL is the current state-of-the-art method for both tasks.
    However, their framework does not consider the pairing relation between each target and opinion, therefore it is difficult to have a completely fair comparison.
    }
    \label{tab:aote}
\end{table}


\section{Analysis}

\subsection{Comparison of Single-word and Multi-word Spans}
\label{ana:single_multi}
We compare the performance of 
Span-ASTE with the previous model GTS \cite{wu-etal-2020-grid} for the following two settings in Table \ref{tab:analysis1}:
Single-Word: Both target and opinion terms in a triplet are single-word spans,
Multi-Word: At least one of the target or opinion terms in a triplet is a multi-word span. 
For the single-word setting, our method shows consistent improvement in terms of both precision and recall score on the four datasets, which results in the improvement of $F_1$ score.
\begin{table*}[!t]
    \centering
    \resizebox{0.9\textwidth}{!}{%
    \begin{tabular}{lllcccccccccccc}
    \toprule
    & \multirow{2}{*}{{\textbf{Mode}}}  & \multirow{2}{*}{{\textbf{Model}}}  & \multicolumn{3}{c}{ \textbf{Rest 14}} & \multicolumn{3}{c}{  \textbf{Lap 14}} & \multicolumn{3}{c}{ \textbf{Rest 15}}  & \multicolumn{3}{c}{  \textbf{Rest 16}}    \\ \cmidrule(lr){4-6} \cmidrule(lr){7-9} \cmidrule(lr){10-12} \cmidrule(lr){13-15}
    &&& $P.$ & $R.$ & $F_1$& $P.$ & $R.$ & $F_1$& $P.$ & $R.$ & $F_1$ & $P.$ & $R.$ & $F_1$ \\
    \midrule
    \multirow{6}{*}{\rotatebox[origin=c]{90}{{BERT}} }& \multirow{3}{*}{{Single-Word}}
    &{GTS}  & 74.93 & 79.15 & 76.98 & 65.47 & 62.54 & 63.97 & 66.55 & 65.66 & 66.10 & 69.66 & 76.74 & 73.03 \\

    & &{Ours} &79.12 & 79.60 & 79.36 & 68.09 & 65.98 & 67.02 & 70.23 & 70.71 & 70.47 & 71.66 & 77.91 & 74.65 \\


    &&{$\Delta$   } & +4.19 & +0.46 & +2.38 & +2.62 & +3.44 & +3.04 & +3.68 & +5.05 & +4.37 & +2.00 & +1.16 & +1.62 \\

    \cmidrule(lr){2-15}
    & \multirow{3}{*}{{Multi-Word}}
    &{GTS} &     56.85 & 49.26 & 52.78 & 52.26 & 41.27 & 46.12 & 50.28 & 47.34 & 48.77 & 56.63 & 55.29 & 55.95 \\
    & &{Ours} & 61.64 & 55.79 & 58.57 & 54.63 & 44.44 & 49.02 & 50.70 & 57.45 & 53.87 & 62.43 & 63.53 & 62.97 \\
    
    &&{$\Delta$   } & 
    
    +4.79 & +6.53 &
    +5.78 & +2.37 &
    +3.17 & +2.90 &
    +0.42 & +10.11 &
    +5.10 & +5.80 &
    +8.24 & +7.02 \\
    
  \bottomrule
    \end{tabular}
    }
    \caption{Analysis with different evaluation modes on the ASTE task.}
    \label{tab:analysis1}
\end{table*}
When we compare the evaluations for multi-word triplets,
our model achieves more significant improvements for $F_1$ scores. Compared to precision, our recall shows greater improvement over the GTS approach.
GTS heavily relies on word-pair interactions to extract triplets,
while our methods explicitly consider the span-to-span interactions. 
Our span enumeration also 
naturally benefits the recall of multi-word spans. 
For both GTS and our model, multi-word triplets pose  challenges and their $F_1$ results drop by more than 10 points, even more than 20 points for Rest 14. 
As shown in Table \ref{tab:dataset}, comparing with the single-word triplets, multi-word triplets are common and account for one-third or even half of the datasets. 
Therefore, a promising direction for future work is to further improve the model's performance on such difficult triplets. 

To identify further areas for improvement,
we analyze the results for the ASTE task based on whether each sentiment triplet contains a multi-word target or multi-word opinion term.
From Table \ref{tab:multi_breakdown}, the results show that the performance is lower when the triplet contains a multi-word opinion term.
This trend can be attributed to the imbalanced data distribution of triplets which contain multi-word target or opinion terms.

\begin{table}[!t]
    \centering
    \resizebox{0.999\columnwidth}{!}{%
    \begin{tabular}{llcccccc}
    \toprule
    \multirow{2}{*}{\textbf{Dataset}}  & \multirow{2}{*}{\textbf{Model}}  & \multicolumn{3}{c}{\textbf{Multi-word Target}} & \multicolumn{3}{c}{\textbf{Multi-word Opinion}}   \\ \cmidrule(lr){3-5} \cmidrule(lr){6-8} 
    && $P.$ & $R.$ & $F_1$&  $P.$ & $R.$ & $F_1$\\
    
    \midrule
    \multirow{2}{*}{Rest 14}
    &{GTS} & 56.54 & 49.81 & 52.96 & 50.67 & 41.76 & 45.78 \\ 
    &{Ours} & 65.96 & 57.62 & \textbf{61.51} & 49.43 & 47.25 & \textbf{48.31} \\
    
    \midrule
    \multirow{2}{*}{Lap 14}
    &{GTS} & 55.11 & 44.09 & 48.99 & 37.50 & 26.09 & \textbf{30.77} \\
    &{Ours} & 56.99 & 48.18 & \textbf{52.22} & 34.62 & 26.09 & 29.75 \\
    
    \midrule
    \multirow{2}{*}{Rest 15}
    &{GTS} & 51.09 & 51.09 & 51.09 & 43.40 & 35.94 & 39.32 \\
    &{Ours} & 55.33 & 60.58 & \textbf{57.84} & 37.18 & 45.31 & \textbf{40.85} \\
    
    \midrule
    \multirow{2}{*}{Rest 16}
    &{GTS} & 62.69 & 65.12 & 63.88 & 28.26 & 24.07 & 26.00 \\
    &{Ours} & 66.43 & 72.09 & \textbf{69.14} & 36.73 & 33.33 & \textbf{34.95} \\
 
    \bottomrule
    \end{tabular}
    }
    
    \caption{
    Further comparison of test results for our model and GTS based on triplets of multi-word targets and opinions for the ASTE task.
    }
    \label{tab:multi_breakdown}
\end{table}

\subsection{Pruning Efficiency}

\begin{figure}
\centering
\resizebox{0.95\linewidth}{!}{
\begin{tikzpicture}
\pgfplotsset{width = 6.5cm, height = 4.5cm}
    \begin{axis}[
        ylabel={$F_1$ (\%)},
        ymax=70,
        ymin=16,
        y label style={yshift=-0.6cm, xshift=0.8cm},
        x label style={yshift=0.4cm, xshift=2.8cm},
        label style={font=\fontsize{7}{1}\selectfont},
        xtick = {1,2,3,4,5,6,7,8,9},
        xticklabels = {0.0625, 0.125, 0.25, 0.50, 1.0},
        xticklabel style = {font=\fontsize{7}{1}\selectfont, rotate=40,},
        yticklabel style = {font=\fontsize{7}{1}\selectfont},
        xtick pos = left,
        ytick pos = left,
        legend pos = south east,
        legend style={font=\fontsize{6.5}{1}\selectfont, row sep=-0.1cm,/tikz/every odd column/.append style={column sep=0.01cm}},
        ymajorgrids = true,
        grid style=dashed,
    ]
    \addplot [mark=square, mark size=1.2pt, color=orange] plot coordinates {
    (1, 55.69) (2, 62.79) (3, 66.47) (4, 67.69) (5, 67.58)};
    \addlegendentry{Dual-Channel};
    \addplot [mark=o,  mark size=1.2pt, color= skyblue] plot coordinates {
    (1, 18.80) (2, 44.51) (3, 60.52) (4, 66.36) (5, 66.82)};
    \addlegendentry{Single-Channel (SC)};
    \addplot [mark=triangle,  mark size=1.2pt, color= gray] plot coordinates {
    (1, 44.51) (2, 60.52) (3, 66.36) (4, 66.82) (5, 67.53)};
    \addlegendentry{SC-Adjusted};
    \end{axis}
\end{tikzpicture}
}
\caption{Dev results with respect to pruning threshold $z$ which intuitively refers to the number of candidate spans to keep per word in the sentence.}
\label{fig:pruning}
\end{figure}
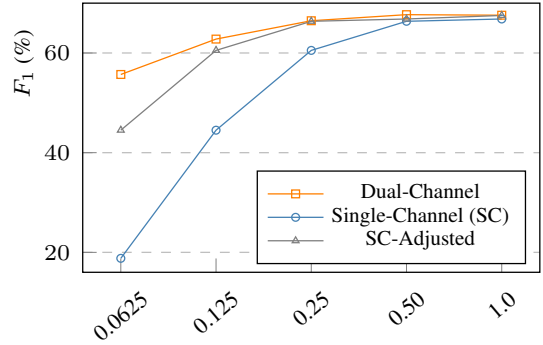

To demonstrate the efficiency of the proposed dual-channel pruning strategy, we also compare it to a simpler strategy, denoted as Single-Channel (SC) which does not distinguish between opinion and target candidates.
Figure \ref{fig:pruning} shows the comparisons.
Note the mention module under this strategy does not explicitly solve the ATE and OTE tasks as it only predicts mention label $m \in \{Valid, Invalid\}$, where $Valid$ means the span is either a target or an opinion span and $Invalid$ means the span does not belong to the two groups.
Given sentence length $n$ and pruning threshold $z$, the number of candidates is limited to $nz$, and hence the computational cost scales with the number of pairwise interactions, $n^2 z^2$.
The dual-channel strategy considers each target-opinion pair where the pruned target and opinion candidate pools both have $nz$ spans. Note that it is possible for the two pools to share some candidates. 
In comparison, the single-channel strategy considers each target-opinion pair where the target and opinion candidates are drawn from the same single pool of $nz$ spans.
In order to consider at least as many target and opinion candidates as the dual-channel strategy, the single-channel strategy has to scale the threshold $z$ by two, which leads to 4 times more pairs and computational cost. 
We denote this setting in Figure \ref{fig:pruning} as SC-Adjusted.
When controlling for computational efficiency, there is a significant performance difference between Dual-Channel and Single-Channel in $F_1$ score, especially for lower values of $z$.
Although the performance gap narrows with increasing $z$, it is not practical for high values.
According to our experimental results, we select the dual-channel pruning strategy with $z=0.5$ for the reported model.

\begin{figure*}[!t]
\centering
\includegraphics[width=0.999\linewidth]{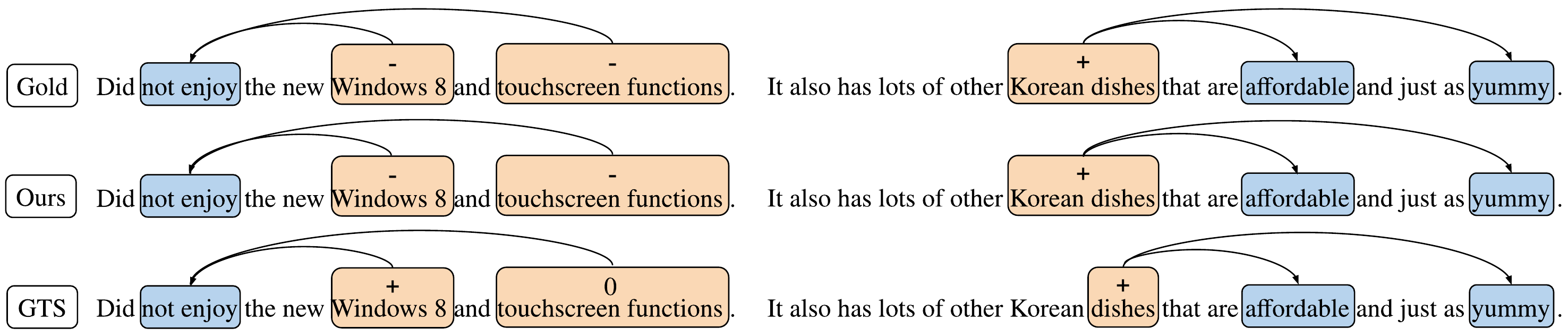}
\caption{
Qualitative analysis.
The target and opinion terms are highlighted in orange and blue respectively.
Each arc indicates the pairing relation between target and opinion terms.
The sentiment polarity of each triplet is indicated above the target terms.
}
\label{fig:case}
\end{figure*}

\subsection{Qualitative Analysis}
\label{sec:case}

To illustrate the differences between the models, we present sample sentences from the ASTE test set
with the gold labels as well as predictions from GTS \cite{wu-etal-2020-grid} and 
Span-ASTE in Figure \ref{fig:case}. 
For the first example, GTS correctly extracts the target term \textit{``Windows 8"} paired with the opinion term \textit{``not enjoy"}, but the sentiment is incorrectly predicted as positive.
When forming the triplet, their decoding heuristic considers the sentiment independently for each word-word pair: 
\{(\textit{``Windows"}, \textit{``not"}, Neutral), (\textit{``8"}, \textit{``not"}, Neutral), (\textit{``Windows"}, \textit{``enjoy"}, Positive), (\textit{``8"}, \textit{``enjoy"}, Positive)\}.
Their heuristic votes the overall sentiment polarity as the most frequent label among the pairs.
In the case of a tie (2 neutral and 2 positive), the heuristic has a predefined bias to assign the sentiment polarity to positive.
Similarly, the word-level method fails to capture the negative sentiment expressed by \textit{``not enjoy"} on the other target term \textit{``touchscreen functions"}.
In the second example, it incompletely extracts the target term \textit{``Korean dishes"}, resulting in the wrong triplet.
For both examples, our method is able to accurately extract the target-opinion pairs and determine the overall sentiment even when each term has multiple words.

\subsection{Ablation Study}

We conduct an ablation study to examine the performance of different modules and span representation methods,
and the results are shown in Table  \ref{tab:ablation}. The average $F_1$ denotes the average dev results of 
Span-ASTE on the four benchmark datasets over 5 runs. Similar to the observation for coreference resolution \cite{lee-etal-2017-end}, we find that the ASTE performance is reduced when removing the span width and distance embedding. 
This indicates that the positional information is still useful for the ASTE task as targets and opinions which are far apart or too long are less likely to form a valid span pair.
As mentioned in Section \ref{representation}, we explore two other methods (i.e., max pooling and mean pooling) to form span representations instead of concatenating the span boundary token representations.
The negative results suggest that using pooling to aggregate the span representation is disadvantageous due to the loss of information that is useful for distinguishing valid and invalid spans.

\begin{table}[t!]
	\centering
	\resizebox{0.999\linewidth}{!}{
		\begin{tabular}{lcc}
			\toprule
			\textbf{Model}& Average $F_1$ & $\Delta F_1$ \\
			\midrule
			Full model  &  \textbf{67.69} & \\
			~~~~W/O width \& distance embedding & 66.45 & -1.24 \\
			~~~~max pooling & 66.09 & -1.60 \\
			~~~~mean pooling & 66.19 & -1.53 \\
			\bottomrule
		\end{tabular}
	}
	\caption{Ablation study on the development sets.}
	\label{tab:ablation}
\end{table}

\section{Related Work}
\label{relatedwork}
Sentiment Analysis is a major Natural Language Understanding (NLU) task \cite{wang-etal-2018-glue} and has been extensively studied as a classification problem at the sentence level \cite{JMLR:v21:20-074, Lan2020ALBERTAL, yang2020xlnet}.
Aspect-Based Sentiment Analysis (ABSA) \cite{pontiki-EtAl:2014:SemEval} 
addresses various sentiment analysis tasks at a fine-grained level. 
As mentioned in the Section \ref{sec:intro}, the subtasks mainly include ASC \cite{dong2014adaptive, zhang2016gated, chen-etal-2017-recurrent, he-etal-2018-effective, tnet2018, peng2018learning, wang2018learning, he-etal-2019-interactive, li-lu-2019-learning, xu2020}, ATE \cite{qiu-etal-2011-opinion, yin2016unsupervised, li2018aspect, ma-etal-2019-exploring}, OTE \cite{minging2004, yang-cardie-2012-extracting, kinger2013, yang-cardie-2013-joint}. There is also another subtask named Target-oriented Opinion Words Extraction (TOWE) \cite{fan2019target}, which
aim to extract the corresponding opinion words for a given target term.
Another line of research focuses on addressing different subtasks together.
Aspect and Opinion Term Co-Extraction (AOTE) aiming to extract the aspect and opinion terms together  \cite{wang2017coupled, ma-etal-2019-exploring, dai2019neural} and is often treated as a sequence labeling problem.
Note that AOTE does not consider the paired sentiment relationship between 
each target and opinion term.
End-to-End ABSA \cite{li2017sentimentscope, ma2018joint, li2019unified,he-etal-2019-interactive}
jointly extracts each aspect term and its associated sentiment
in an end-to-end manner. 
A few other methods are recently proposed to jointly solve three or more subtasks of ABSA. 
\citet{chen-qian-2020-relation} proposed a relation aware collaborative learning framework to unify the three fundamental subtasks and achieved strong performance on each subtask and combined task.
While \citet{Wan_Yang_Du_Liu_Qi_Pan_2020} focused more on aspect category related subtasks, such as Aspect Category Extraction and Aspect Category and Target Joint Extraction.
ASTE \cite{peng2019knowing,wu-etal-2020-grid,Xu2020PositionAwareTF,zhang-etal-2020-multi-task} is the most recent development of ABSA and its aim is to extract and form the aspect term, its associated sentiment, and the corresponding opinion term into a triplet.

\section{Conclusions}
In this work, we propose a span-level approach - Span-ASTE to learn the interactions between target spans and opinion spans for the ASTE task.
It can address the limitation of the existing approaches that only consider word-to-word interactions.
We also propose to include the ATE and OTE tasks as supervision for our dual-channel pruning strategy to reduce the number of enumerated target and opinion candidates to increase the computational efficiency and maximize the chances of pairing valid target and opinion candidates together.
Our method significantly outperforms the previous 
methods for ASTE as well as ATE and OTE tasks and our analysis demonstrates the effectiveness of our approach.
While we achieve strong performance on the ASTE task,
the performance can be mostly attributed to the improvement on the multi-word triplets.
As discussed in Section \ref{ana:single_multi}, there is still a significant 
performance gap between single-word and multi-word triplets,
and this 
can be a potential area for future work.

\bibliographystyle{acl_natbib}
\bibliography{acl2021}

\appendix

\section{Additional Experimental Settings} 
We run our model experiments on a Nvidia Tesla V100 GPU, with CUDA version 10.2 and PyTorch version 1.6.0. 
The average run time for BERT-based model is 157 sec/epoch, 115 sec/epoch, 87 sec/epoch, and 111 sec/epoch for Rest 14, Lap 14, Rest 15, and Rest 16 respectively.
The total number of parameters is 2.24M when GloVe is used, and is 110M when BERT base is used.
The feed-forward neural networks in the mention module and triplet module have 2 hidden layers and hidden size of 150.
We use ReLU activation and dropout of 0.4 after each hidden layer.
We use Xavier Normal weight initialization for the feed-forward parameters.
The span width and distance embeddings have 20 and 128 dimensions respectively.
Their input values are bucketed \cite{Gardner2017AllenNLP}   before being fed to an embedding matrix lookup: [0, 1, 2, 3, 4, 5-7, 8-15, 16-31, 32-63, 64+].
During training, the model parameters are updated after each sentence which results in a batch size of 1.
For each input text sequence, we restrict it to a maximum of 512 tokens.

\section{Additional Data Statistics}
Table \ref{tab:adddataset} shows the number of target terms and opinion terms on the four datasets.

\section{Dev Results}
Table \ref{tab:dev} shows the results of our model on the development datasets.

\begin{table}[!t]
    \centering
    \resizebox{0.999\columnwidth}{!}{%
    \begin{tabular}{llcccccc}
    \toprule
    \multirow{2}{*}{\textbf{Dataset}}  & \multirow{2}{*}{\textbf{Model}}  & \multicolumn{3}{c}{\textbf{ATE}} & \multicolumn{3}{c}{\textbf{OTE}}   \\ \cmidrule(lr){3-5} \cmidrule(lr){6-8} 
    && $P.$ & $R.$ & $F_1$&  $P.$ & $R.$ & $F_1$\\
    \midrule
    \multirow{3}{*}{Rest 14}
    & {JET}$^o_{M=6}$ & 83.21 & 66.04 & 73.64 & 83.76 & 77.28 & 80.39 \\
    & {GTS} & 83.25 & 81.49 & 82.36 & 86.55 & 86.65 & \textbf{86.60} \\
    & {Ours} & 86.20 & 80.31 & \textbf{83.15} & 87.20 & 84.54 & 85.85 \\
    
    \midrule
    \multirow{3}{*}{Lap 14}
    & {JET}$^o_{M=6}$ & 83.33 & 68.03 & 74.91 & 77.16 & 75.53 & 76.34 \\
    & {GTS} & 82.17 & 73.65 & 77.68 & 81.63 & 74.05 & 77.66 \\
    & {Ours} & 87.69 & 75.38 & \textbf{81.07} & 85.61 & 76.58 & \textbf{80.84} \\

    \midrule
    \multirow{3}{*}{Rest 15}
    & {JET}$^o_{M=6}$ & 83.04 & 65.74 & 73.38 & 81.33 & 68.98 & 74.65 \\
    & {GTS} & 80.95 & 74.77 & 77.74 & 80.96 & 76.57 & 78.70 \\
    & {Ours} & 81.60 & 78.01 & \textbf{79.76} & 80.09 & 81.13 & \textbf{80.61} \\

    \midrule
    \multirow{3}{*}{Rest 16}
    & {JET}$^o_{M=6}$ & 83.33 & 68.58 & 75.24 & 89.44 & 80.21 & 84.57 \\
    & {GTS} & 82.69 & 85.62 & 84.13 & 83.37 & 86.53 & 84.92 \\
    & {Ours} & 84.20 & 86.06 & \textbf{85.12} & 84.62 & 88.00 & \textbf{86.28} \\

  \bottomrule
    \end{tabular}
    }
    \caption{
    Test results on the ATE and OTE tasks with sub-optimal evaluation method.
    Our method and GTS \cite{wu-etal-2020-grid} allow for ATE and OTE tasks to be predicted independently from the ASTE task.
    However, JET$^o_{M=6}$ \cite{Xu2020PositionAwareTF} does not.
    Hence, we make another comparison here by extracting all opinion and target spans from the ASTE predictions.
    }
    \label{tab:jet}
\end{table}

\begin{table}[!t]
    \centering
    \resizebox{0.999\columnwidth}{!}{%
    \begin{tabular}{llcccccc}
    \toprule
    \multirow{2}{*}{\textbf{Dataset}}  & \multirow{2}{*}{\textbf{Model}}  & \multicolumn{3}{c}{\textbf{ATE}} & \multicolumn{3}{c}{\textbf{OTE}}   \\ \cmidrule(lr){3-5} \cmidrule(lr){6-8} 
    && $P.$ & $R.$ & $F_1$&  $P.$ & $R.$ & $F_1$\\
    \midrule
    \multirow{3}{*}{Rest 14}
    &{GTS} & 78.50 & 87.38 & 82.70 & 82.07 & 88.99 & 85.39 \\ 
    &{RACL} & 79.90 & 87.74 & 83.63 & 80.26 & 87.99 & 83.94 \\
    &{Ours} & 83.56 & 87.59 & \textbf{85.50} & 82.93 & 89.67 & \textbf{86.16} \\
    
    \midrule
    \multirow{3}{*}{Lap 14}
    &{GTS} & 78.63 & 81.86 & 80.21 & 76.27 & 79.32 & 77.77 \\ 
    &{RACL} & 78.11 & 81.99 & 79.99 & 75.12 & 79.92 & 77.43 \\
    &{Ours} &  81.48 & 86.39 & \textbf{83.86} & 83.00 & 82.28 & \textbf{82.63 }\\
    
    \midrule
    \multirow{3}{*}{Rest 15}
    &{GTS} & 74.95 & 82.41 & 78.50 & 74.75 & 81.56 & 78.01 \\
    &{RACL} & 75.22 & 81.94 & 78.43 & 76.41 & 82.56 & 79.35 \\
    &{Ours} &  78.97 & 84.68 & \textbf{81.72} & 77.36 & 84.86 & \textbf{80.93}\\
    
    \midrule
    \multirow{3}{*}{Rest 16}
    &{GTS} & 75.05 & 89.16 & 81.50 & 78.36 & 88.42 & 83.09 \\ 
    &{RACL} & 74.12 & 89.20 & 80.95 & 79.25 & 89.77 & 84.17 \\
    &{Ours} &  79.78 & 88.50 & \textbf{83.89} & 82.59 & 90.91 & \textbf{86.54} \\
 
  \bottomrule
    \end{tabular}
    }
    \caption{
    Additional comparison of test results on the ATE and OTE tasks. 
    Note that RACL \cite{chen-qian-2020-relation} does not consider supervision from target-opinion pairs, but it includes the sentiment polarities on the target terms.
    }
    \label{tab:racl}
\end{table}

\begin{table*}[t]
\centering
\resizebox{0.85\textwidth}{!}{
\begin{tabular}{lcccccccc}
\toprule 
\multirow{2}{*}{\textbf{Dataset}} & \multicolumn{2}{c}{\textbf{Rest 14}} & \multicolumn{2}{c}{\textbf{Lap 14}} & \multicolumn{2}{c}{\textbf{Rest 15}} & \multicolumn{2}{c}{\textbf{Rest 16}} \\ 
\cmidrule(lr){2-3}\cmidrule(lr){4-5}\cmidrule(lr){6-7}\cmidrule(lr){8-9}
& \# Target & \# Opinion & \# Target & \# Opinion & \# Target & \# Opinion & \# Target & \# Opinion \\
\midrule
\textbf{Train} & 2051 & 2086 & 1281 & 1268 & {\color{white}1}862 & {\color{white}1}941 & 1198 & 1307 \\
\textbf{Dev} & {\color{white}1}500 & {\color{white}1}503 & {\color{white}1}296 & {\color{white}1}304 & {\color{white}1}213 & {\color{white}1}236 & {\color{white}1}296 & {\color{white}1}319 \\
\textbf{Test} & {\color{white}1}848 & {\color{white}1}854 & {\color{white}1}463 & {\color{white}1}474 & {\color{white}1}432 & {\color{white}1}461 & {\color{white}1}452 & {\color{white}1}475 \\
\bottomrule
\end{tabular}
}
\caption{Additional statistics. \# Target denotes the number of target terms.
\# Opinion denotes the numbers of opinion terms.
}
\label{tab:adddataset}
\end{table*}

\begin{table*}[!t]
    \centering
    \resizebox{0.99\textwidth}{!}{%
    \begin{tabular}{lcccccccccccc}
    \toprule
    \multirow{2}{*}{\textbf{Model}}  & \multicolumn{3}{c}{ Rest 14} & \multicolumn{3}{c}{  Lap 14} & \multicolumn{3}{c}{ Rest 15}  & \multicolumn{3}{c}{  Rest 16}    \\ 
    \cmidrule(lr){2-4} \cmidrule(lr){5-7} \cmidrule(lr){8-10} \cmidrule(lr){11-13} 
    & $P.$ & $R.$ & $F_1$& $P.$ & $R.$ & $F_1$& $P.$ & $R.$ & $F_1$ & $P.$ & $R.$ & $F_1$ \\
    \midrule

    {Ours (BiLSTM)} &  
    66.76 & 53.90 & 59.61 & 60.78 & 49.37 & 54.45 & 69.13 & 60.08 & 64.26 & 71.59 & 61.95 & 66.41 \\
    {Ours (BERT)} &  
    68.05 & 65.65 & 66.80 & 63.35 & 58.90 & 61.02 & 70.16 & 71.41 & 70.75 & 72.52 & 71.92 & 72.19 \\
  \bottomrule
    \end{tabular}
    }
    \caption{Results on the development datasets.}
    \label{tab:dev}
\end{table*}

\section{Additional Comparisons}
As mentioned by footnote 5 in Section 3.5, we cannot make a direct comparison with the JET model \cite{Xu2020PositionAwareTF}, as it is not able to directly solve the ATE and OTE tasks unless the evaluation is conducted based on the triplet results. Table \ref{tab:jet} shows such comparisons. Our proposed method generally outperforms the previous two end-to-end approaches on the four datasets.

As mentioned in Table 3, it is challenging to make a fair comparison between the previous ABSA framework RACL \cite{chen-qian-2020-relation}, which also address the ATE and OTE tasks while solving other ABSA subtasks, and our approach as well as the GTS \cite{wu-etal-2020-grid}. This is because the mentioned approaches have different task settings. The RACL considers the sentiment polarity on the target terms when solving the ATE and OTE tasks, but GTS and our method both consider the pairing relation between target and opinion terms. However, for reference, Table \ref{tab:racl} shows the comparisons of the three methods on the ATE and OTE tasks on the datasets released by \citet{Xu2020PositionAwareTF}.

\end{document}